\documentclass{article}

\PassOptionsToPackage{numbers, compress}{natbib}

\usepackage[preprint]{neurips_2024}

\usepackage{graphicx}
\usepackage{subfigure}
\usepackage{multirow}
\usepackage{amsmath}
\usepackage{amssymb}
\usepackage{mathtools}
\usepackage{amsthm}
\usepackage{algorithm}
\usepackage{algorithmic}
\usepackage[utf8]{inputenc} 
\usepackage[T1]{fontenc}    
\usepackage{hyperref}       
\usepackage{url}            
\usepackage{booktabs}       
\usepackage{amsfonts}       
\usepackage{nicefrac}       
\usepackage{microtype}      
\usepackage{xcolor}         
\definecolor{yes}{rgb}{0.22, 0.11, 1.0}
\definecolor{no}{rgb}{1.0, 0.48, 0.36}
\definecolor{na}{rgb}{0.55, 0.52, 0.49}

\title{Understanding Robust Overfitting from the Feature Generalization Perspective}

\author{%
  Chaojian Yu \\
  The University of Sydney\\
  \texttt{chyu8051@uni.sydney.edu.au} \\
  \And
  Xiaolong Shi \\
  The University of Sydney \\
  \texttt{xiaolong3166@gmail.com} \\
  \AND
  Jun Yu \\
  University of Science and Technology of China \\
  \texttt{harryjun@ustc.edu.cn} \\
  \And
  Bo Han \\
  Hong Kong Baptist University \\
  \texttt{bhanml@comp.hkbu.edu.hk} \\
  \And
  Tongliang Liu \\
  The University of Sydney \\
  \texttt{tongliang.liu@uni.sydney.edu.au} \\
}

\begin{document}

\maketitle

\begin{abstract}
  Adversarial training (AT) constructs robust neural networks by incorporating adversarial perturbations into natural data. However, it is plagued by the issue of robust overfitting (RO), which severely damages the model's robustness. In this paper, we investigate RO from a novel feature generalization perspective. Specifically, we design factor ablation experiments to assess the respective impacts of natural data and adversarial perturbations on RO, identifying that the inducing factor of RO stems from natural data. Given that the only difference between adversarial and natural training lies in the inclusion of adversarial perturbations, we further hypothesize that adversarial perturbations degrade the generalization of features in natural data and verify this hypothesis through extensive experiments. Based on these findings, we provide a holistic view of RO from the feature generalization perspective and explain various empirical behaviors associated with RO. To examine our feature generalization perspective, we devise two representative methods, attack strength and data augmentation, to prevent the feature generalization degradation during AT. Extensive experiments conducted on benchmark datasets demonstrate that the proposed methods can effectively mitigate RO and enhance adversarial robustness.
\end{abstract}

\section{Introduction}
\label{Introduction}
Adversarial training (AT)~\citep{madry2018towards} has been proved to be reliable to improve a model’s robustness against adversarial attacks~\citep{szegedy2014intriguing,goodfellow2015explaining}. It constructs robust neural networks by incorporating adversarial perturbations into natural data and training models using adversarial data generated on-the-fly. AT has been established as one of the most effective empirical defenses~\citep{athalye2018obfuscated}. Although promising to improve the model’s robustness, AT is plagued by the issue of robust overfitting (RO)~\citep{rice2020overfitting}. That is, after a certain point during AT, the model’s robust accuracy continues to substantially decrease with further training. This phenomenon has been consistently observed across different datasets, network architectures, and AT variants~\citep{rice2020overfitting}.

In this work, we investigate RO from a novel feature generalization perspective. Firstly, we conduct a series of factor ablation experiments to identify the inducing factors of RO. Specifically, we consider natural data and adversarial perturbations from small-loss training data as distinct factors and employ factor ablation adversarial training to assess their individual impacts on RO. We observe that simultaneously removing adversarial perturbations and natural data from small-loss training data greatly mitigate RO. However, experiments that only remove adversarial perturbations still exhibits a severe degree of RO. Given that these factor ablation experiments strictly adhere to the controlling variables principle, with the sole variation being the inclusion of natural data in the training set, we can infer that the inducing factor of RO stems from natural data.

Given that the only difference between adversarial and natural training lies in the inclusion of adversarial perturbations in the natural data, we further hypothesize that adversarial perturbations may degrade the generalization of features in natural data and lead to the emergence of RO. To validate this hypothesis, we introduce an additional adversarial perturbation into the training natural data during AT. We observe that as the additional adversarial perturbation is incorporated, the model’s test robustness deteriorates. This suggests that the additional adversarial perturbations have a detrimental effect on feature generalization. Furthermore, we are able to simulate the RO phenomenon during AT by linearly increasing the budget of additional adversarial perturbations. These experimental results demonstrate that adversarial perturbations can degrade feature generalization and induce the emergence of RO phenomenon.

Based on these findings, we provide a holistic view of RO from the feature generalization perspective. In standard AT, a robustness gap exists between the training and test data due to factors like the memorization effect of deep networks~\cite{arpit2017closer, dong2022exploring} and the distribution deviation between the finite training and test data. Certain small-loss training data maintain a significant robustness gap when compared to the test data. However, adversarial perturbations are generated on the fly and adaptively adjusted based on the model's robustness. The robustness gap between training and test data leads to distinct adversarial perturbations for the features in natural data. These distinct adversarial perturbations degrade feature generalization. The degradation of feature generalization, in turn, weakens the model’s test performance and widens the robustness gap between training and test data, thus forming a vicious cycle. Additionally, we provide explanations for various empirical behaviors associated with RO based on our feature generalization perspective.

To examine our feature generalization perspective, we propose two representative methods: attack strength and data augmentation. Specifically, the attack strength method adjusts adversarial perturbations on small-loss training data to verify their impact on feature generalization, and the data augmentation method adjusts model's robustness on small-loss training data to assess the effect of the robustness gap on feature generalization. Both methods can effectively prevent feature generalization degradation during AT and mitigate RO, thereby validating our feature generalization understanding of RO. Additionally, extensive experiments conducted on benchmark datasets demonstrate the efficacy of these methods in enhancing adversarial robustness. To sum up, our contributions are as follows:
\begin{itemize}
\item We conduct factor ablation AT adhering to the principle of controlling variables and identify that the inducing factor of RO stems from natural data.
\item We further hypothesize that adversarial perturbations degrade the generalization of features in natural data and validate this hypothesis through extensive experiments.
\item Based on these findings, we provide a holistic view of RO from the feature generalization perspective and explain various empirical behaviors associated with RO.
\item To examine our feature generalization perspective, we propose two representative methods to prevent the degradation of feature generalization. Extensive experiments demonstrate that the proposed methods effectively mitigate RO and enhance adversarial robustness.
\end{itemize}

\section{Related Work}
\label{Related Work}
In this section, we provide a brief overview of related works from two perspectives: AT and RO.

\subsection{Adversarial Training}
Let $f_\theta$, $\mathcal{X}$ and $\ell$ denote the network $f$ with model parameter $\theta$, the input space, and the loss function, respectively. Given a $C$-class dataset $\mathcal{S}=\{(x_i,y_i)\}^{n}_{i=1}$, where $x_i \in \mathcal{X}$ and $y_i \in \mathcal{Y} = \{0,1,\dots,C-1\}$ denotes its corresponding label, the objective function of natural training is defined as follows:
\begin{equation}
\min_\theta \frac{1}{n}\sum_{i=1}^{n} \ell(f_\theta(x_{i}),y_{i}),
\label{standard training}
\end{equation}
where the network $f_\theta$ learns features in natural data that are correlated with associated labels to minimize the empirical risk of misclassifying natural inputs. However, evidence~\citep{szegedy2014intriguing,tsipras2018robustness,ilyas2019adversarial} suggests that naturally trained models tend to learn fragile features that are vulnerable to adversarial attacks. To address this issue, AT~\citep{goodfellow2015explaining,madry2018towards} introduces adversarial perturbations to each data point by transforming $ \mathcal{S}=\{(x_i,y_i)\}^{n}_{i=1}$ into $\mathcal{S}'=\{(x_i'=x_i+\delta_i,y_i)\}^{n}_{i=1}$. The adversarial perturbations $\{\delta_i\}^{n}_{i=1}$ are constrained by a pre-specified budget, \textit{i.e.} $\{\delta \in \Delta: ||\delta||_p \leq \epsilon\}$, where $p$ can be $1,2,\infty$, etc. Therefore, the objective of AT is to solve the following min-max optimization problem:
\begin{equation}
\min_\theta \frac{1}{n}\sum_{i=1}^{n} \max_{\delta_i \in \Delta} \ell(f_\theta(x_{i}+\delta_i),y_{i}),
\label{adversarial training}
\end{equation}
where the inner maximization process aims to generate adversarial data that maximizes the classification loss, and the outer minimization process optimizes model parameter using the generated adversarial data. This iterative procedure aims to train an adversarially robust classifier. The most commonly employed approach for generating adversarial data is Projected Gradient Descent (PGD)~\citep{madry2018towards}, which applies adversarial attack to natural data $x_i$ over multiple steps $k$ with a step size of $\alpha$:
\begin{equation}
\delta^{k} = \Pi_{\Delta}(\alpha \cdot \mathrm{sign}\nabla_{x}\ell(f_\theta(x+\delta^{k-1}),y) + \delta^{k-1}), k \in \mathbb{N},
\label{PGD}
\end{equation}
where $\delta^{k}$ represents the adversarial perturbation at step $k$, and $\Pi_{\Delta}$ denotes the projection operator.

Besides standard AT, there are also several AT variants~\citep{kannan2018adversarial,zhang2019theoretically,wang2019improving}. One typical variant is TRADES~\citep{zhang2019theoretically}, which trains the network on both natural data and adversarial data:
\begin{equation}
    \min_\theta \sum_{i} \big\{  \mathrm{CE}(f_{\theta}(x_i),y_i) + \beta \cdot \max_{\delta_i \in \Delta} \mathrm{KL}(f_{\theta}(x_i)||f_{\theta}(x_{i}+\delta_i)) \big\},
\end{equation}
where CE is the cross-entropy loss that encourages the network to maximize natural accuracy, KL is the Kullback-Leibler divergence that encourages robustness improvement, and the hyperparameter $\beta$ is used to regulate the tradeoff between natural accuracy and adversarial robustness.

Moreover, further improvements to adversarial training have also been explored, such as curriculum adversarial training~\citep{cai2018curriculum,wang2019convergence}, ensemble adversarial training~\citep{tramer2018ensemble,pang2019improving}, and adaptive budget schedule~\citep{liu2020loss,addepalli2022scaling,jia2022adversarial}.

\subsection{Robust Overfitting}
RO was initially observed in standard AT~\citep{madry2018towards}. Subsequent systematic studies by Rice et al.~\citep{rice2020overfitting} showed that traditional remedies for overfitting in deep learning offer limited help in addressing RO in AT, prompting further efforts to explore RO. Schmidt et al.~\citep{schmidt2018adversarially} attributed RO to sample complexity theory and suggested that robust generalization requires more training data, which is supported by empirical results in derivative works~\citep{carmon2019unlabeled,alayrac2019labels,zhai2019adversarially}. Other works also proposed various strategies to mitigate RO without relying on additional training data, such as sample reweighting~\citep{wang2019improving,zhang2020geometry, liu2021probabilistic}, label smoothing~\citep{izmailov2018averaging}, stochastic weight averaging~\citep{chen2020robust}, temporal ensembling~\citep{dong2022exploring}, knowledge distillation~\citep{chen2020robust}, weight regularization~\citep{wu2020adversarial,yu2022understanding}, and data augmentation~\citep{rebuffi2021data,tack2022consistency,li2023data}. Additionally, RO has been studied from various perspectives, such as label noise~\citep{dong2022label}, input-loss landscape ~\citep{li2023understanding} and minimax game~\citep{wang2023balance}. In this work, we provide a novel feature generalization understanding of RO.

\section{Method}
\label{Method}
In this section, we begin with factor ablation experiments to identify the inducing factors of RO (Section~\ref{S3.1}). Furthermore, we propose a hypothesis that adversarial perturbations degrade feature generalization and verify this hypothesis through extensive experiments. Following this, we provide a holistic view of RO from the feature generalization perspective and explain various empirical behaviors associated with RO (Section~\ref{S3.2}). Lastly, we present two representative methods aimed at examining our feature generalization understanding of RO (Section~\ref{S3.3}).

\begin{figure}[t]
\centering
    \subfigure[Factor ablation AT]{\includegraphics[width=0.24\columnwidth]{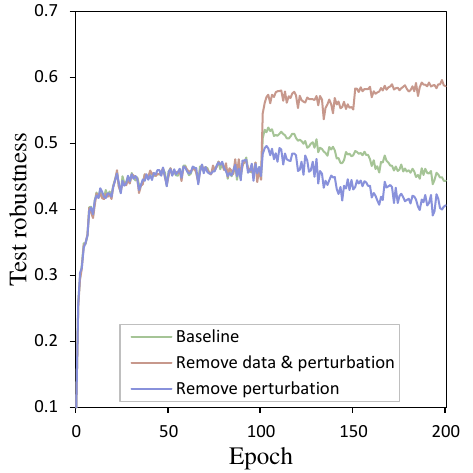}}
    \subfigure[Verification]{\includegraphics[width=0.24\columnwidth]{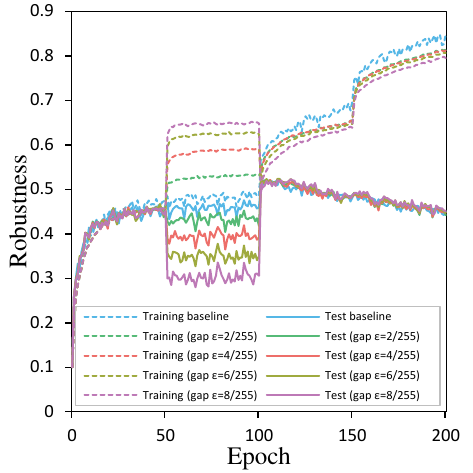}}
    \subfigure[Inducing RO]{\includegraphics[width=0.24\columnwidth]{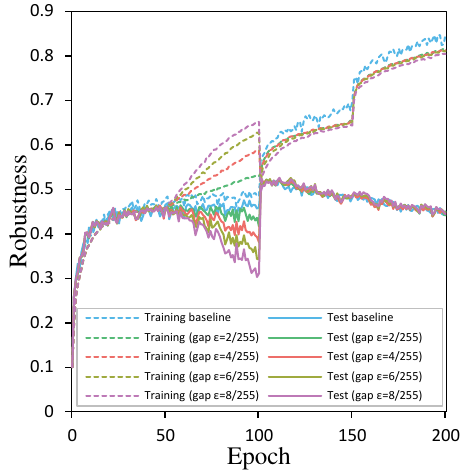}}
    \subfigure[Standard AT]{\includegraphics[width=0.24\columnwidth]{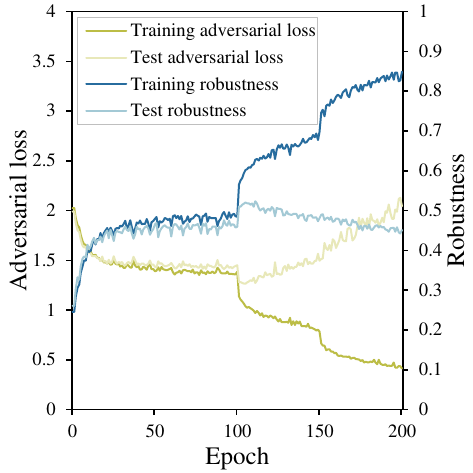}}
\caption{(a) Test robustness of factor ablation AT; (b) Verification experiments with varying budgets of additional adversarial perturbations, (c) Inducing RO with linearly increasing budgets of additional adversarial perturbations, and (d) Adversarial loss and robustness of standard AT.}
\label{fig:1}
\end{figure}

\subsection{Factor Ablation Adversarial Training}
\label{S3.1} 
Drawing inspiration from the data ablation experiments conducted by Yu et al. \cite{yu2022understanding}, which investigated the influence of small-loss training data on RO by excluding such data during training, we propose factor ablation adversarial training as a method to pinpoint the inducing factors of RO. Following a similar experimental setup, we also utilize a fixed loss threshold to differentiate small-loss training data. For instance, in the CIFAR10 dataset, data with an adversarial loss below 1.7 is regarded as small-loss training data. Unlike data ablation experiments that treat adversarial data as a single entity, our approach considers natural data and adversarial perturbations from small-loss training data as separate factors. We conduct more granular factor ablation adversarial training to identify the specific factor inducing RO. Specifically, we train a PreAct ResNet-18 model on CIFAR-10 dataset under the $\ell_\infty$ threat model and remove designated factors before RO occurs (i.e., at the 100th epoch). These experiments include: \textit{\romannumeral1)} \textbf{baseline}, which is a baseline group where no factors are removed; \textit{\romannumeral2)} \textbf{data \& perturbation}, which removes both natural data and adversarial perturbations from small-loss training data; and \textit{\romannumeral3)} \textbf{perturbation}, which solely removes adversarial perturbations from small-loss training data.

It's important to note that these experiments strictly adhere to the principle of controlling variables. They are entirely identical before RO occurred, with the sole distinction among different experimental groups being the removal of designated factors from the training set. The experimental results of factor ablation adversarial training are summarized in Figure~\ref{fig:1}(a). It is observed that the \textbf{data \& perturbation} group demonstrates a significant alleviation of RO, whereas both the \textbf{baseline} and \textbf{perturbation} groups continue to exhibit severe RO. Given that the only difference between the \textbf{data \& perturbation} and \textbf{perturbation} groups lies in the inclusion of natural data in the training set, we can infer that the inducing factor of RO stems from natural data. Similar effects can be observed across different datasets, network architectures, and AT variants (see Appendix~\ref{A.A}), indicating that it is a general finding in adversarial training.

\subsection{Analysis of Robust Overfitting from the Feature Generalization Perspective}
\label{S3.2}
Based on the factor ablation experiments in Section~\ref{S3.1}, we know that natural data acts as the inducing factor of RO. It’s noteworthy that natural data is also present in natural training. However, in natural training, RO phenomenon is typically absent. For instance, deep learning models trained on natural data often achieve zero training error, effectively memorizing the training data, without observable detrimental effects on generalization performance \cite{rice2020overfitting}. In contrast, RO is a dominant phenomenon in AT, which commonly exhibits significantly degraded generalization performance. Given that the only difference between adversarial and natural training lies in the inclusion of adversarial perturbations in the natural data, a natural hypothesis arises: adversarial perturbations may degrade the generalization of features in natural data and lead to the emergence of RO.

\begin{figure}[t]
  \centering 
  \includegraphics[width=0.45\textwidth]{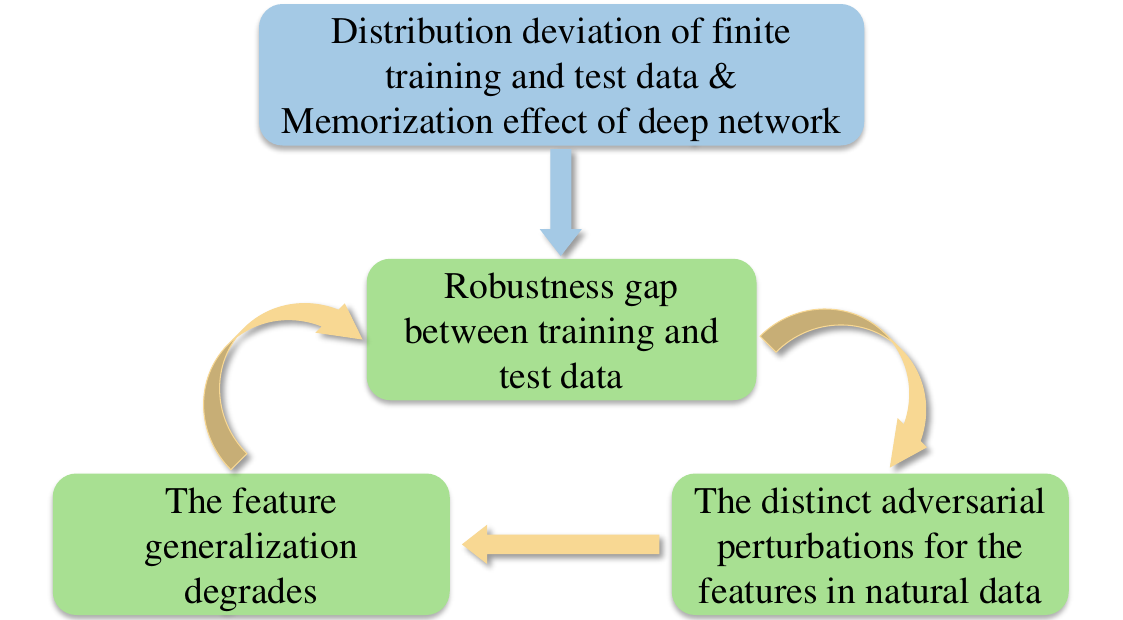}
  \caption{
  Illustration of the analysis of RO from the feature generalization perspective. In standard AT, a robustness gap exists between the training and test data due to factors like the memorization effect of deep networks and the distribution deviation between the finite training and test data. Subsequently, the robustness gap between training and test data leads to distinct adversarial perturbations for the features in natural data. These distinct adversarial perturbations degrade the feature generalization. The degradation of feature generalization further widens the model’s robustness gap between training and test data, thus forming a vicious cycle.
  }
  \label{fig:fig}
\end{figure}

\textbf{Verification.} To validate the aforementioned hypothesis, before generating training adversarial data in AT, we introduce an additional adversarial perturbation into the training natural data and observe the model's robust generalization performance. Specifically, before the onset of RO (i.e., between the 50th and 100th epochs), we incorporate an extra adversarial perturbation in the opposite direction into the training natural data and observe the model’s test robustness. We conduct experiments with varying budgets of additional adversarial perturbations, such as $\epsilon=\{0/255, 2/255, 4/255, 6/255, 8/255\}$. The model's test robustness is shown in Figure~\ref{fig:1}(b). It is important to note that the additional adversarial perturbations are introduced only on the training natural data, while the model and the test data remain unchanged, ensuring that the model’s robustness on the test data remains unaffected. However, we observe that as the additional adversarial perturbation is incorporated, the model’s test robustness deteriorates, indicating that the additional adversarial perturbations exert a detrimental effect on feature generalization. Furthermore, we linearly increase the budget of additional adversarial perturbations between the 50th and 100th epochs, and the model’s test robustness is summarized in Figure~\ref{fig:1}(c). It is observed that the RO phenomenon can be simulated by adjusting the adversarial perturbations on training data. In Appendix~\ref{B.B}, we also provide more experimental results conducted at different stages of training. These experimental results support our hypothesis that the adversarial perturbations can degrade feature generalization and induce the emergence of RO.

\textbf{A holistic view of RO from the feature generalization perspective.}
In standard AT, there exists a robustness gap between the training and test data before the onset of RO, as shown in Figure~\ref{fig:1}(d). This robustness gap arises due to factors such as the memorization effect of deep networks~\cite{arpit2017closer,dong2022exploring} and the distribution deviation between the finite training and test data. Certain small-loss training data maintain a significant robustness gap when compared to the test data.
However, adversarial perturbations are generated on the fly and adaptively adjusted based on the model’s robustness. For instance, consider a single data point, when the model’s robustness is poor, adversarial perturbations mainly impede the model’s learning of non-robust features. Conversely, when the model’s robustness is strong, adversarial perturbation will hinder the model’s learning of robust features because the process of generating adversarial perturbation aims to maximize the classification loss. The robustness gap between training and test data leads to distinct adversarial perturbations for the features in natural data, degrading feature generalization. As feature generalization deteriorates, the model's adversarial robustness on the test data progressively declines, and the robustness gap between training and test data continues to widen, forming a vicious cycle, as illustrated in Figure~\ref{fig:fig}.

Based on the analysis above, some small-loss training data maintain a significant robustness gap compared to the test data. This can explain why removing small-loss adversarial data from the training data can effectively prevent RO. Furthermore, AT exhibits some other empirical behaviors associated with RO, for instance, as the perturbation budget increases, the extent of RO initially rises and then decreases. Our feature generalization perspective provides an intuitive explanation: as the perturbation budget increases from 0, the impact of adversarial perturbations on the generalization of features in natural data also grows, leading to a higher probability of RO phenomenon during AT. This can explain why natural training does not exhibit RO, and as the adversarial perturbation budget increases, the extent of RO also rises. However, with a further increase in the perturbation budget, the model's robustness on the training data is limited, leading to a narrowing robustness gap between training and test data. The narrowing robustness gap reduces the generation of adversarial perturbations that degrade feature generalization. Therefore, the extent of RO gradually decreases.

\subsection{The Proposed Methods}
\label{S3.3}
In Section~\ref{S3.2}, we analyze RO as a result of the degradation of Feature Generalization (ROFG), which arises from the distinct adversarial perturbations for the features in natural data. In this part, we introduce two approaches to examine our feature generalization understanding of RO.

\textbf{ROFG through attack strength.} 
In order to verify our analysis that adversarial perturbations on some small-loss training data degrade feature generalization, we conduct AT with varying attack strengths to generate adversarial perturbations on the small-loss training data before RO occurs (i.e., at the 100th epoch). Specifically, we use a fixed loss threshold $t=1.7$ to distinguish small-loss training data and train PreAct ResNet-18 on CIFAR10 under the $\ell_\infty$ threat model with adjusted perturbation budgets $\epsilon_a$ on small-loss training data, ranging from $0/255$ to $24/255$. The pseudocode is provided in Algorithm~\ref{alg:1}, which utilizes attack strength to adjust the adversarial perturbations on small-loss training data, referred to as ROFG$_\mathrm{AS}$.

\begin{figure}[t]
\centering
    \subfigure[ROFG$_\mathrm{AS}$]{\includegraphics[width=0.24\columnwidth]{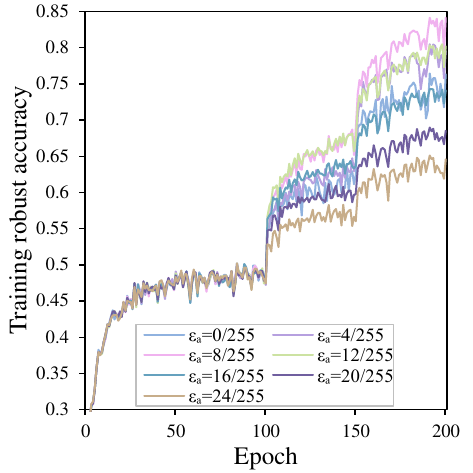}
    \includegraphics[width=0.24\columnwidth]{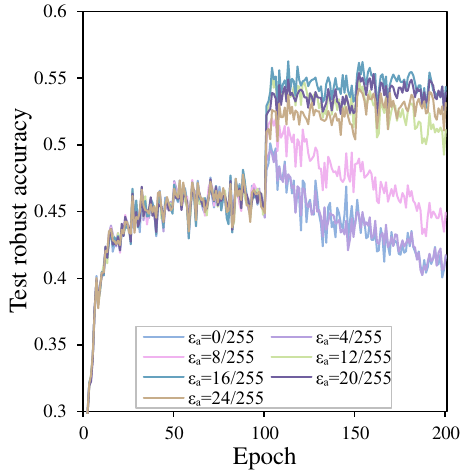}
    }
    \subfigure[ROFG$_\mathrm{DA}$]{\includegraphics[width=0.24\columnwidth]{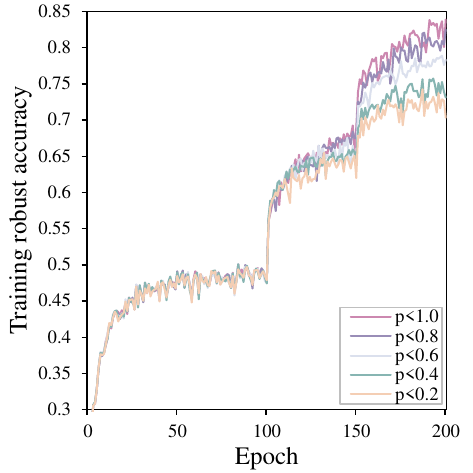}
    \includegraphics[width=0.24\columnwidth]{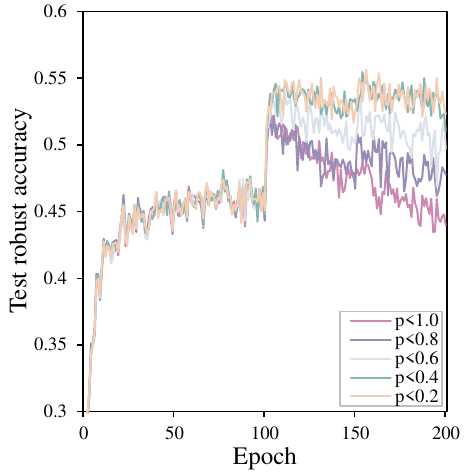}
    }
\caption{(a) The learning curves of ROFG$_\mathrm{AS}$ with varying attack strengths, and (b) the learning curves of ROFG$_\mathrm{DA}$ with different proportions of small-loss training data.}
\label{fig:method}
\end{figure}

For ROFG$_\mathrm{AS}$ with different perturbation budgets $\epsilon_a$, we evaluate the model's training and test robustness under the standard perturbation budget of $\epsilon=8/255$, with the learning curves summarized in Figure~\ref{fig:method}(a). We observe a clear correlation between the applied perturbation budgets $\epsilon_a$ and the extent of RO. Specifically, the greater the perturbation budgets $\epsilon_a$, the milder the extent of RO. When the perturbation budget $\epsilon_a$ exceeds a certain threshold, such as $16/255$, the model exhibits almost no RO. It is worth noting that similar patterns are also observed across different datasets, network architectures, and AT variants (as shown in Appendix~\ref{D.D}). These experimental results demonstrate that adjusting adversarial perturbations on small-loss training data can effectively prevent the degradation of feature generalization, thereby alleviating the RO phenomenon, validating our feature generalization understanding of RO.

\begin{algorithm}[t]
\small
   \caption{ROFG$_\mathrm{AS}$}
   \label{alg:1}
\begin{algorithmic}
   \STATE {\bfseries Input:} Network $f_\theta$, training data $\mathcal{S}$, mini-batch $\mathcal{B}$, batch size $n$, learning rate $\eta$, PGD step size $\alpha$, PGD budget $\epsilon$, PGD steps $K$, adjusted PGD budget $\epsilon_a$, adjusted PGD steps $K_a$, small-loss threshold $t$.
   \STATE {\bfseries Output:} Adversarially robust model $f_\theta$.
   \REPEAT
   \STATE Read mini-batch $(x_{\mathcal{B}},y_{\mathcal{B}})$ from training set $\mathcal{S}$.
   \STATE $x_{\mathcal{B}}' \leftarrow x_{\mathcal{B}} + \delta$, where $\delta \sim \mathrm{Uniform}(-\epsilon,\epsilon)$
   \FOR{$k=1$ {\bfseries to} $K$}
   \STATE $x_{\mathcal{B}}' \leftarrow \Pi_{\epsilon}(x_{\mathcal{B}}'+\alpha \cdot \mathrm{sign}(\nabla_{x_{\mathcal{B}}'}\ell(f_\theta(x_{\mathcal{B}}'),y_{\mathcal{B}})))$
   \ENDFOR
   \STATE $x_t =x_{\mathcal{B}}(\ell(f_\theta(x_{\mathcal{B}}'),y_{\mathcal{B}})\le t)$  \hfill \#Distinguish small-loss training data $x_t$
   \STATE $y_t =y_{\mathcal{B}}(\ell(f_\theta(x_{\mathcal{B}}'),y_{\mathcal{B}})\le t)$
   \STATE $x_t' \leftarrow x_t + \delta$, where $\delta \sim \mathrm{Uniform}(-\epsilon_a,\epsilon_a)$
   \FOR{$k=1$ {\bfseries to} $K_a$}
   \STATE $x_t' \leftarrow \Pi_{\epsilon_a}(x_t'+\alpha \cdot \mathrm{sign}(\nabla_{x_t'}\ell(f_\theta(x_t'),y_t)))$ \hfill \#Generate adjusted adversarial perturbations for $x_t$
   \ENDFOR
   \STATE $x_{\mathcal{B}}'(\ell(f_\theta(x_{\mathcal{B}}'),y_{\mathcal{B}}) \le t) = x_t'$ \hfill \#Replace the adversarial perturbations of $x_t$
   \STATE $\theta \leftarrow \theta - \eta\nabla_\theta\frac{1}{n}\sum_{i=1}^{n}\ell(f_\theta({x'}_{\mathcal{B}}^{(i)}),y_{\mathcal{B}}^{(i)})$
   \UNTIL{training converged}
\end{algorithmic}
\end{algorithm}

\begin{algorithm}[t]
\small
   \caption{ROFG$_\mathrm{DA}$}
   \label{alg:2}
\begin{algorithmic}
   \STATE {\bfseries Input:} Network $f_\theta$, training data $\mathcal{S}$, mini-batch $\mathcal{B}$, batch size $n$, learning rate $\eta$, PGD step size $\alpha$, PGD budget $\epsilon$, PGD steps $K$, small-loss data proportion $p$, small-loss threshold $t$.
   \STATE {\bfseries Output:} Adversarially robust model $f_\theta$.
   \REPEAT
   \STATE Read mini-batch $(x_{\mathcal{B}},y_{\mathcal{B}})$ from training set $\mathcal{S}$.
   \STATE $x_{\mathcal{B}}' \leftarrow x_{\mathcal{B}} + \delta$, where $\delta \sim \mathrm{Uniform}(-\epsilon,\epsilon)$
   \FOR{$k=1$ {\bfseries to} $K$}
   \STATE $x_{\mathcal{B}}' \leftarrow \Pi_{\epsilon}(x_{\mathcal{B}}'+\alpha \cdot \mathrm{sign}(\nabla_{x_{\mathcal{B}}'}\ell(f_\theta(x_{\mathcal{B}}'),y_{\mathcal{B}})))$
   \ENDFOR
   \STATE $n_t=\sum{\mathbb{I}}(\ell(f_\theta(x_{\mathcal{B}}'),y_{\mathcal{B}})\le t)$ \hfill \#Count the number of small-loss training data
   \IF{$n_t/n > p$}
   \REPEAT
   \STATE $x_t = x_{\mathcal{B}}(\ell(f_\theta(x_{\mathcal{B}}'),y_{\mathcal{B}}) \le t)$ \hfill \#Distinguish small-loss training data $x_t$
   \STATE $y_t =y_{\mathcal{B}}(\ell(f_\theta(x_{\mathcal{B}}'),y_{\mathcal{B}}) \le t)$
   \STATE $x_t = \mathrm{DataAugmentation}(x_t)$ \hfill \#Conduct data augmentation on $x_t$
   \STATE $x_t' \leftarrow x_t + \delta$, where $\delta \sim \mathrm{Uniform}(-\epsilon,\epsilon)$
   \FOR{$k=1$ {\bfseries to} $K$}
   \STATE $x_t' \leftarrow \Pi_{\epsilon}(x_t'+\alpha \cdot \mathrm{sign}(\nabla_{x_t'}\ell(f_\theta(x_t'),y_t)))$
   \ENDFOR
   \STATE $x_{\mathcal{B}}'(\ell(f_\theta(x_{\mathcal{B}}'),y_{\mathcal{B}})\le t)(\ell(f_\theta(x_t'),y_t)>t) = x_t'(\ell(f_\theta(x_t'),y_t)>t)$ \hfill \#Update training data
   \STATE $n_t=\sum{\mathbb{I}}(\ell(f_\theta(x_{\mathcal{B}}'),y_{\mathcal{B}})\le t)$ \hfill \#Update the number of small-loss training data
   \UNTIL{$n_t/n \le p$}
   \ENDIF
   \STATE $\theta \leftarrow \theta - \eta\nabla_\theta\frac{1}{n}\sum_{i=1}^{n}\ell(f_\theta({x'}_{\mathcal{B}}^{(i)}),y_{\mathcal{B}}^{(i)})$
   \UNTIL{training converged}
\end{algorithmic}
\end{algorithm}

\textbf{ROFG through data augmentation.}
To further support our analysis that the model’s robustness gap between training and test data promotes the generation of adversarial perturbations that degrades feature generalization, we employ data augmentation techniques to weaken the model's robustness on small-loss training data before RO occurs (i.e., at the 100th epoch). However, the image transformations in data augmentation techniques are often stochastic and may not yield the desired effect. To address this issue, we iteratively apply random image transformations to the small-loss training data, consistently selecting the transformed data that does not fall within the small-loss scope. Specifically, we use a fixed loss threshold $t=1.7$ to distinguish small-loss training data, and set a threshold $p$ for the proportion of small-loss training data when train PreAct ResNet-18 on CIFAR10 under the $\ell_\infty$ threat model. At the beginning of each iteration, we check whether the proportion of small-loss training data meets the specified threshold $p$. If this proportion is below the specified threshold $p$, we apply random image transformations to the small-loss training data until the specified threshold $p$ is reached.
The pseudocode is provided in Algorithm~\ref{alg:2}, where the adopted data augmentation technique is AugMix~\citep{hendrycks2020augmix}. We refer to this adversarial training pipeline, empowered by the data augmentation technique, as ROFG$_\mathrm{DA}$.

The learning curves of ROFG$_\mathrm{DA}$ with different thresholds $p$ are summarized in Figure~\ref{fig:method}(b). We observe a clear correlation between the threshold $p$ and the extent of RO. As the robustness gap between small-loss training data and test data decreases, the extent of RO becomes increasingly mild. Furthermore, the observed effects are consistent across different datasets, network architectures, and AT variants (as shown in Appendix~\ref{E.E}). These experimental results show that narrowing the model’s robustness gap between training and test data can effectively prevent the degradation of feature generalization, supporting our feature generalization understanding of RO.

\section{Experiment}
\label{Experiment}
In this section, we validate the effectiveness of the proposed methods. In Section~\ref{S4.1}, we provide the ablation analysis of ROFG$_\mathrm{AS}$ and ROFG$_\mathrm{DA}$. Section~\ref{S4.2} presents the performance evaluation of ROFG$_\mathrm{AS}$ and ROFG$_\mathrm{DA}$. In Section~\ref{S4.3}, we discuss the limitations of the proposed methods.

\textbf{Setup.}
We conducted extensive experiments across different benchmark datasets (CIFAR10 and CIFAR100~\citep{krizhevsky2009learning}), network architectures (PreAct ResNet-18~\citep{he2016deep} and Wide ResNet-34-10~\citep{zagoruyko2016wide}), and adversarial training approaches (AT~\citep{madry2018towards}, TRADES~\citep{zhang2019theoretically}, AWP~\citep{wu2020adversarial}, and MLCAT~\citep{yu2022understanding}). For training, we follow the same experimental settings as outlined in Rice et al.~\citep{rice2020overfitting}. Regarding robustness evaluation, we employ PGD-20~\citep{madry2018towards} and AutoAttack (AA)~\citep{croce2020reliable} as adversarial attack methods. Detailed descriptions of the experimental setup can be found in Appendix~\ref{F.F}.

\begin{figure}[t]
\centering
    \subfigure[Impact of attack strength]{
        \includegraphics[width=0.185\columnwidth]{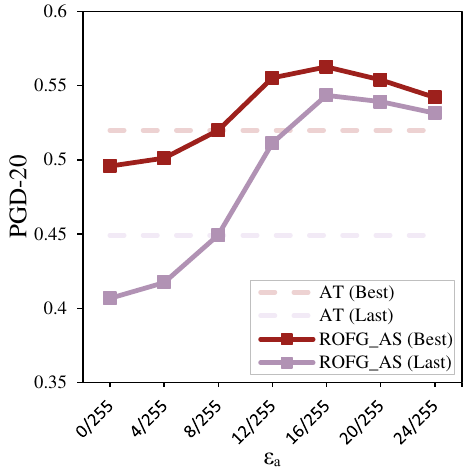}
        \includegraphics[width=0.185\columnwidth]{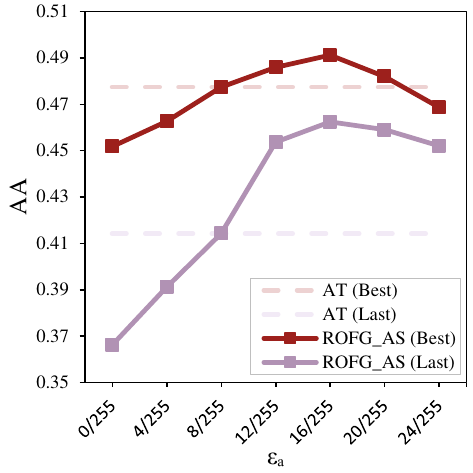}
    }
    \subfigure[Importance of loss threshold $t$]{
        \includegraphics[width=0.185\columnwidth]{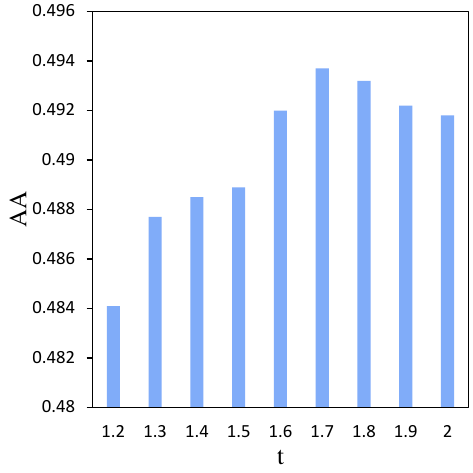}
    }
    \subfigure[Effect of data augmentation]{
        \includegraphics[width=0.185\columnwidth]{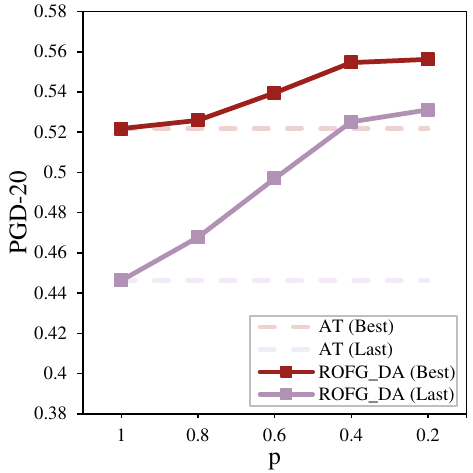}
        \includegraphics[width=0.185\columnwidth]{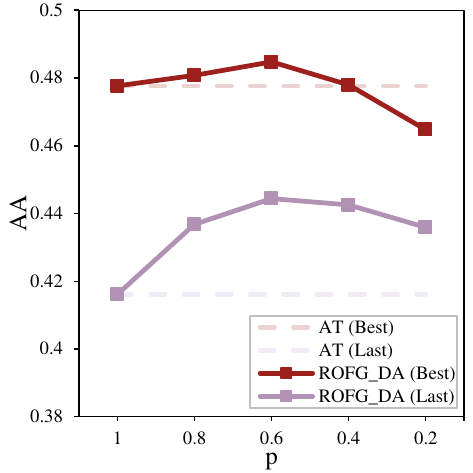}
    }
\caption{Ablation analysis of ROFG$_\mathrm{AS}$ and ROFG$_\mathrm{DA}$.}
\vspace{-0.2cm}
\label{fig:ablation analysis}
\end{figure}

\subsection{Analysis of the Proposed Methods}
\label{S4.1}
We investigate the impacts of algorithmic components in ROFG$_\mathrm{AS}$ and ROFG$_\mathrm{DA}$ under the PreAct ResNet-18 architecture on the CIFAR10 dataset.

\textbf{The impact of attack strength.} 
We empirically verify the effectiveness of attack strength component in ROFG$_\mathrm{AS}$ by comparing the performances of models trained using different perturbation budgets $\epsilon_a$. Specifically, we conduct ROFG$_\mathrm{AS}$ with the perturbation budget $\epsilon_a$ ranging from 0/255 to 24/255, and the results are summarized in Figure~\ref{fig:ablation analysis}(a). It is evident that as $\epsilon_a$ increases, the robustness gap between the ``Best'' and ``Last'' checkpoints steadily decreases, demonstrating its effectiveness. For adversarial robustness, we observe a trend of initially increasing and then decreasing. This is not surprising since the attack strength component with excessively large perturbation budgets is inherently detrimental to robustness improvement under the standard perturbation budget ($\epsilon=8/255$). Therefore, in practice, we need to adjust the perturbation budget $\epsilon_a$ to balance the mitigation of RO and robustness improvement.

\textbf{The importance of loss threshold $t$.} Identifying exactly which adversarial perturbations degrade feature generalization is generally intractable. To address this, we introduce a fixed loss threshold $t$, which is used to distinguish small-loss training data. We conduct ROFG$_\mathrm{AS}$ experiments with $t$ varying from 1.2 to 2.0, and the model’s robustness performances are summarized in Figure~\ref{fig:ablation analysis}(b). We observe a trend of initially increasing and then decreasing robustness, which implies that adversarial perturbations degrading feature generalization are primarily concentrated in the small-loss training data, suggesting the importance of the loss threshold $t$.

\textbf{The effect of data augmentation.} We further investigate the effect of the data augmentation component by conducting ROFG$_\mathrm{DA}$ with different thresholds $p$. The results are summarized in Figure~\ref{fig:ablation analysis}(c). It is observed that as $p$ decreases, the robustness gap between the "Best" and "Last" checkpoints steadily shrinks. Additionally, we explore the use of different data augmentation techniques to implement ROFG$_\mathrm{DA}$ in Appendix~\ref{G.G}. It is observed that all implementations of ROFG$_\mathrm{DA}$ can effectively mitigate RO, suggesting that the proposed approach is generally effective regardless of the chosen data augmentation technique. For adversarial robustness, as $p$ decreases, the model’s robustness performance initially increases and then decreases. This trend may be due to the data augmentation component weakening the model’s robustness on the training data, thereby degrading the model’s defense against strong attacks.

\begin{table*}[t]
  \small
  \centering
  \caption{Test robustbess (\%) of ROFG$_\mathrm{AS}$ and ROFG$_\mathrm{DA}$ across different datasets and adversarial training methods based on the PreAct ResNet-18 architecture.}
  \label{table:1}
\resizebox{0.99\linewidth}{!}{
  \begin{tabular}{llcccccccc}
    \toprule
    \multirow{2}*{Method} & \multirow{2}*{Dataset} & \multicolumn{2}{c}{Natural} && \multicolumn{2}{c}{PGD-20} && \multicolumn{2}{c}{AA}\\
    \cmidrule{3-4}
    \cmidrule{6-7}
    \cmidrule{9-10}
    & & Best & Last && Best & Last && Best & Last \\
    \midrule
    AT & \multirow{3}*{CIFAR10} & 82.31$\pm$0.18 & 84.11$\pm$0.40 && 52.28$\pm$0.37 & 44.46$\pm$0.14 && 48.09$\pm$0.11 & 42.01$\pm$0.10 \\
    ROFG$_{\mathrm{DA}}$ & & \textbf{82.58$\pm$0.25} & \textbf{85.45$\pm$0.24} && 53.95$\pm$0.27 & 49.69$\pm$0.10 && 48.48$\pm$0.09 & 44.44$\pm$0.18\\
    ROFG$_{\mathrm{AS}}$ &  & 77.68$\pm$0.52 & 78.04$\pm$0.10 && \textbf{56.37$\pm$0.33} & \textbf{51.96$\pm$0.08} && \textbf{49.37$\pm$0.19} & \textbf{45.97$\pm$0.16} \\
    \midrule
    AT & \multirow{3}*{CIFAR100} & 55.14$\pm$0.12 & 55.83$\pm$0.20 && 28.93$\pm$0.36 & 20.87$\pm$0.33 && 24.53$\pm$0.02 & 18.92$\pm$0.20 \\
    ROFG$_{\mathrm{DA}}$ &  & \textbf{55.79$\pm$0.39} & \textbf{57.92$\pm$0.50} && 29.40$\pm$0.15 & 25.51$\pm$0.17 && 24.80$\pm$0.12 & 21.59$\pm$0.03 \\
    ROFG$_{\mathrm{AS}}$ &  & 51.02$\pm$0.27 & 51.06$\pm$0.35 && \textbf{30.25$\pm$0.36} & \textbf{26.19$\pm$0.31} && \textbf{25.63$\pm$0.05} & \textbf{22.67$\pm$0.14}\\
    \midrule
    TRADES & \multirow{3}*{CIFAR10} & 81.50$\pm$0.31 & 82.27$\pm$0.40 && 52.92$\pm$0.19 & 49.95$\pm$0.21 && 48.90$\pm$0.16 & 46.92$\pm$0.09\\
    TRADES-ROFG$_{\mathrm{DA}}$ &  & \textbf{82.89$\pm$0.25} & \textbf{83.28$\pm$0.17} && 53.14$\pm$0.24 & \textbf{52.13$\pm$0.36} && 49.12$\pm$0.13 & 48.41$\pm$0.06 \\
    TRADES-ROFG$_{\mathrm{AS}}$ & & 80.92$\pm$0.52 & 80.97$\pm$0.52 && \textbf{53.49$\pm$0.31} & 52.04$\pm$0.25 && \textbf{49.88$\pm$0.15} & \textbf{48.91$\pm$0.07}\\
    \midrule
    AWP & \multirow{3}*{CIFAR10} & 81.01$\pm$0.52 & 81.61$\pm$0.46 && 55.36$\pm$0.36 & 55.05$\pm$0.13 && 50.12$\pm$0.07 & 49.85$\pm$0.20\\
    AWP-ROFG$_{\mathrm{DA}}$ & & \textbf{81.12$\pm$0.30} & \textbf{81.63$\pm$0.44} && 55.89$\pm$0.22 & 55.32$\pm$0.18 && 50.49$\pm$0.13 & 50.19$\pm$0.13 \\
    AWP-ROFG$_{\mathrm{AS}}$ & & 79.39$\pm$0.14 & 79.85$\pm$0.41 && \textbf{56.06$\pm$0.11} & \textbf{55.67$\pm$0.20} && \textbf{50.79$\pm$0.10} & \textbf{50.45$\pm$0.08}\\
    \midrule
    MLCAT & \multirow{3}*{CIFAR10} & 81.70$\pm$0.19 & 82.26$\pm$0.29 && 58.33$\pm$0.26 & 58.25$\pm$0.12 && 50.54$\pm$0.10 & 50.46$\pm$0.11\\
    MLCAT-ROFG$_{\mathrm{DA}}$ & & \textbf{82.06$\pm$0.13} & \textbf{82.50$\pm$0.20} && 58.76$\pm$0.37 & 58.57$\pm$0.22 && 50.61$\pm$0.16 & 50.52$\pm$0.08 \\
    MLCAT-ROFG$_{\mathrm{AS}}$ & & 80.64$\pm$0.37 & 81.41$\pm$0.17 && \textbf{58.91$\pm$0.38} & \textbf{58.74$\pm$0.17} && \textbf{50.88$\pm$0.10} & \textbf{50.65$\pm$0.06}\\
    \bottomrule
  \end{tabular}
  }
\end{table*}

\begin{table*}[t]
  \small
  \centering
  \caption{Test robustbess (\%) of ROFG$_\mathrm{AS}$ across different datasets and adversarial training methods based on the Wide ResNet-34-10 architecture.}
  \label{table:2}
\resizebox{0.99\linewidth}{!}{
  \begin{tabular}{llcccccccc}
    \toprule
    \multirow{2}*{Method} & \multirow{2}*{Dataset} & \multicolumn{2}{c}{Natural} & & \multicolumn{2}{c}{PGD-20} & & \multicolumn{2}{c}{AA}\\
    \cmidrule{3-4}
    \cmidrule{6-7}
    \cmidrule{9-10}
    & & Best & Last && Best & Last && Best & Last \\
    \midrule
    AT & \multirow{2}*{CIFAR10} & \textbf{85.49$\pm$0.23} & \textbf{86.50$\pm$0.33} && 55.40$\pm$0.37 & 47.14$\pm$0.29 && 52.31$\pm$0.13 & 45.74$\pm$0.16 \\
    ROFG$_{\mathrm{AS}}$ &  & 82.64$\pm$0.12 & 82.71$\pm$0.55 && \textbf{59.07$\pm$0.21} & \textbf{51.18$\pm$0.19} && \textbf{53.04$\pm$0.08} & \textbf{47.22$\pm$0.19}\\
    \midrule
    AT & \multirow{2}*{CIFAR100} & \textbf{60.90$\pm$0.13} & \textbf{59.07$\pm$0.31} && 31.35$\pm$0.25 & 26.03$\pm$0.32 && 27.42$\pm$0.07 & 24.39$\pm$0.07 \\
    ROFG$_{\mathrm{AS}}$ & & 55.78$\pm$0.32 & 52.75$\pm$0.40 && \textbf{33.48$\pm$0.26} & \textbf{28.40$\pm$0.25} && \textbf{28.86$\pm$0.02} & \textbf{24.58$\pm$0.18}\\
    \midrule
    TRADES & \multirow{2}*{CIFAR10} & \textbf{84.78$\pm$0.51} & \textbf{84.70$\pm$0.13} && 56.25$\pm$0.37 & 48.49$\pm$0.21 && 53.12$\pm$0.14 & 46.69$\pm$0.15\\
    TRADES-ROFG$_{\mathrm{AS}}$ & & 83.36$\pm$0.48 & 83.17$\pm$0.42 && \textbf{57.07$\pm$0.23} & \textbf{49.65$\pm$0.20} && \textbf{53.79$\pm$0.10} & \textbf{47.71$\pm$0.04}\\
    \midrule
    AWP & \multirow{2}*{CIFAR10} & \textbf{85.30$\pm$0.12} & \textbf{85.39$\pm$0.29} && 58.35$\pm$0.36 & 57.16$\pm$0.29 && 53.07$\pm$0.09 & 52.49$\pm$0.04\\
    AWP-ROFG$_{\mathrm{AS}}$ & & 84.40$\pm$0.22 & 84.71$\pm$0.50 && \textbf{59.56$\pm$0.09} & \textbf{57.65$\pm$0.21} && \textbf{54.70$\pm$0.12} & \textbf{53.08$\pm$0.17}\\
    \midrule
    MLCAT & \multirow{2}*{CIFAR10} & \textbf{86.72$\pm$0.16} & \textbf{87.32$\pm$0.15} && 62.63$\pm$0.27 & 61.91$\pm$0.24 && 54.73$\pm$0.10 & 54.61$\pm$0.13\\
    MLCAT-ROFG$_{\mathrm{AS}}$ &  & 86.46$\pm$0.43 & 86.74$\pm$0.18 && \textbf{63.59$\pm$0.32} & \textbf{62.40$\pm$0.11} && \textbf{55.38$\pm$0.12} & \textbf{54.85$\pm$0.05}\\
    \bottomrule
  \end{tabular}
  }
\end{table*}

\subsection{Performance Evaluation}
\label{S4.2}
The proposed ROFG$_\mathrm{AS}$ and ROFG$_\mathrm{DA}$ methods can effectively mitigate RO, as demonstrated in Section~\ref{S3.3}, Appendix~\ref{D.D}, and Appendix~\ref{E.E}, validating our feature generalization perspective. Regarding adversarial robustness, we conduct experiments under the PreAct ResNet-18 architecture across different datasets and adversarial training methods. The performance of the ``Best'' and ``Last'' checkpoints is provided in Table~\ref{table:1}. It is observed that for methods exhibiting RO, such as AT~\citep{madry2018towards} and TRADES~\citep{zhang2019theoretically}, the proposed methods significantly improve model robustness. For methods already mitigating RO, such as AWP~\citep{wu2020adversarial} and MLCAT~\citep{yu2022understanding}, our methods still can achieve some complementary robustness improvement, demonstrating the significance of our feature generalization perspective. In Table~\ref{table:1}, we can see that ROFG$_\mathrm{AS}$ outperforms ROFG$_\mathrm{DA}$ under AA in all settings. Thus we further conduct experiments using ROFG$_\mathrm{AS}$ under the Wide ResNet-34-10 architecture, with the results shown in Table~\ref{table:2}. It is observed that under the stronger backbone network, the ROFG$_\mathrm{AS}$ method significantly enhances model robustness across different datasets and adversarial training methods. Additionally, we conduct additional experiments using ROFG$_\mathrm{AS}$ on the Tiny-ImageNet dataset~\citep{tinyimagenet} and VGG-16 architecture~\citep{simonyan2014very}, with the results shown in Appendix~\ref{H.H}. Consistently, the proposed approach mitigates RO and boosts adversarial robustness, demonstrating its effectiveness.

\subsection{Limitations}
\label{S4.3}
While the proposed methods contribute to validating our feature generalization perspective, they have certain limitations. Firstly, they introduce additional components into the standard AT pipeline, thereby increasing computational complexity. For instance, under a single GeForce RTX 3080 GPU, the average training time per epoch in standard AT is 147.7s, while for ROFG$_\mathrm{AS}$ and ROFG$_\mathrm{DA}$, it extends to 261.6s and 1222.9s, respectively. Moreover, our methods have yet to achieve state-of-the-art adversarial robustness, and they often cannot simultaneously achieve optimal robustness improvement and RO mitigation, as discussed in Section~\ref{S4.1}. However, it is worth noting that the main focus of this work is to provide a novel feature generalization understanding of RO. These methods are specifically tailored to verify our feature generalization perspective, and these limitations essentially do not compromise our understanding of RO.

\section{Conclusion}
\label{Conclusion}
This paper investigate RO from a novel feature generalization perspective. First, we conduct factor ablation adversarial training and identify that the inducing factor of RO stems from natural data. Furthermore, we hypothesize adversarial perturbations degrade the generalization of features in natural data and validate this hypothesis through extensive experiments. Based on these findings, we provide a holistic view of RO from the feature generalization perspective and explain various empirical behaviors associated with RO. We also design two representative methods to examine our feature generalization perspective: attack strength and data augmentation. Extensive experiments demonstrate that the proposed methods effectively mitigate RO and can enhance adversarial robustness across different adversarial training methods, network architectures, and benchmark datasets.


\bibliographystyle{abbrv}
\bibliography{reference.bib}


\newpage
\appendix

\section{Additional Results for Factor Ablation Adversarial Training}
\label{A.A}
\begin{figure}[t]
\centering
    \subfigure[]{
        \includegraphics[width=0.31\columnwidth]{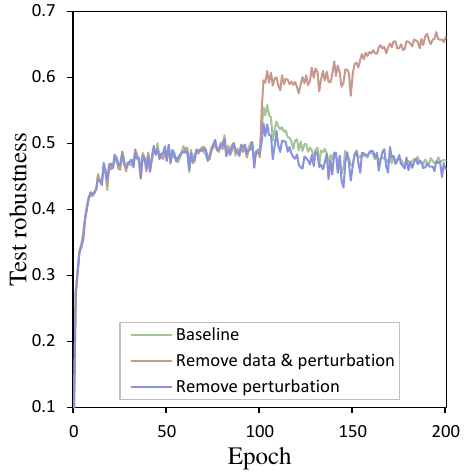}
    }
    \subfigure[]{
        \includegraphics[width=0.31\columnwidth]{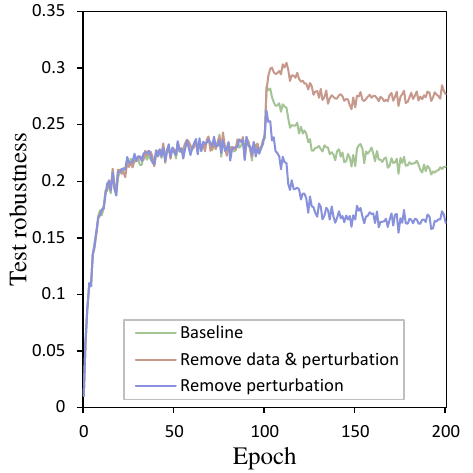}
    }
    \subfigure[]{
        \includegraphics[width=0.31\columnwidth]{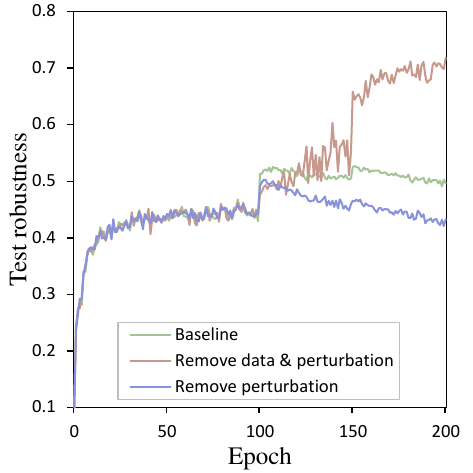}
    }
\caption{Experimental results of factor ablation adversarial training (a) on CIFAR10 dataset using Wide ResNet-34-10 with AT, (b) on CIFAR100 dataset using PreAct ResNet-18 with AT, and (c) on CIFAR10 dataset using PreAct ResNet-18 with TRADES.}
\label{fig:3}
\end{figure}

In this section, we present additional results regarding factor ablation adversarial training across different datasets, network architectures, and AT variants. We utilize the same experimental setup as described in Section~\ref{S3.1}, employing a fixed loss threshold of 1.7 for the CIFAR10 dataset and 4.0 for the CIFAR100 dataset to distinguish small-loss training data. In the \textbf{data \& perturbation} group, we remove both natural data and adversarial perturbations from small-loss training data. In the \textbf{perturbation} group, we solely remove adversarial perturbations from small-loss training data. The experimental results are summarized in Figure~\ref{fig:3}. It is observed that the \textbf{data \& perturbation} group consistently exhibits significant relief from RO, while the \textbf{perturbation} group shows severe RO. Since the only difference between the \textbf{data \& perturbation} and \textbf{perturbation} groups is the inclusion of natural data in the training set, these experimental results suggest that the inducing factor of RO stems from natural data.

\begin{figure}[t]
\centering
    \subfigure[]{
        \includegraphics[width=0.23\columnwidth]{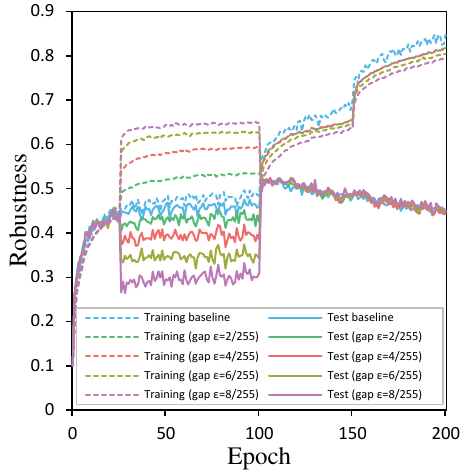}
    }
    \subfigure[]{
        \includegraphics[width=0.23\columnwidth]{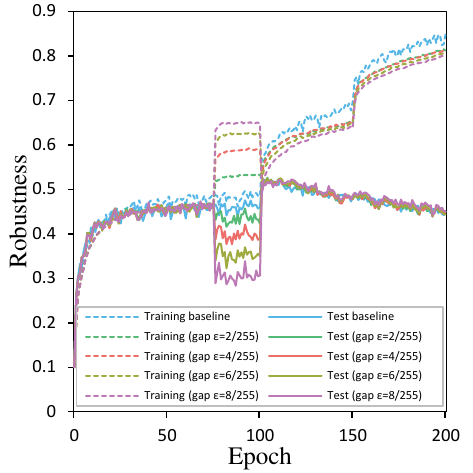}
    }
    \subfigure[]{
        \includegraphics[width=0.23\columnwidth]{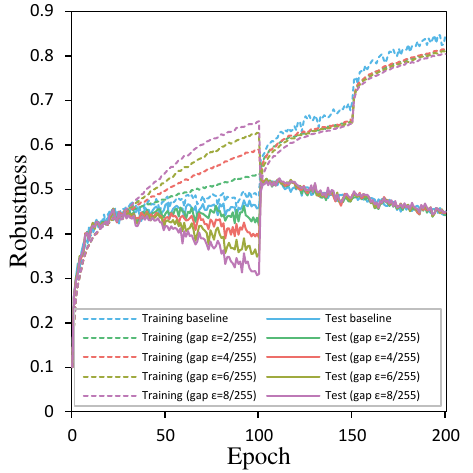}
    }
    \subfigure[]{
        \includegraphics[width=0.23\columnwidth]{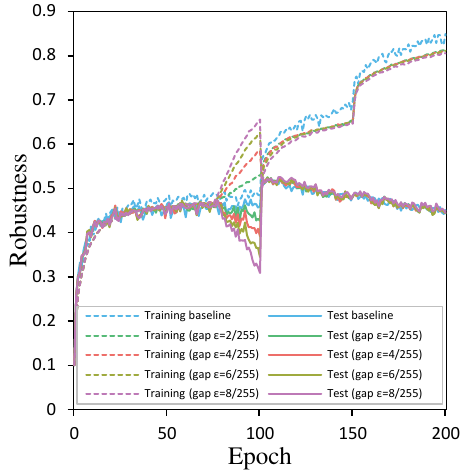}
    }
\caption{Verification experiments with varying budgets of additional adversarial perturbations (a) between the 25th and 100th epochs, and (b) between the 75th and 100th epochs. Inducing experiments with linearly increasing budgets of additional adversarial perturbations (c) between the 25th and 100th epochs, and (d) between the 75th and 100th epochs.}
\label{fig:BB}
\end{figure}

\section{Additional Results for Verification and Inducing Experiments}
\label{B.B}
In this section, we present additional results from verification and inducing experiments conducted at different stages of training. We utilize the same experimental setup as described in Section~\ref{S3.2}, which involves incorporating an additional adversarial perturbation in the opposite direction into the training natural data and observing the model’s test robustness. Specifically, we conduct experiments with varying budgets of additional adversarial perturbations at the 25th and 75th epochs, such as $\epsilon=\{0/255, 2/255, 4/255, 6/255, 8/255\}$. The model's test robustness is shown in Figure~\ref{fig:BB}(a) and (b). It is observed that as the additional adversarial perturbation is incorporated, the model’s test robustness deteriorates at different stages of training, indicating that the additional adversarial perturbations exert a detrimental effect on feature generalization.
It is worth noting that if we replace the additional adversarial perturbations with random noise of the same budget, the model does not exhibit the corresponding robust generalization degradation. This indicates that it is not taken for granted that adding extra perturbations to the training natural data will degrade the model’s test robustness.
Furthermore, we linearly increase the budget of additional adversarial perturbations at the 25th and 75th epochs, and the model’s test robustness is summarized in Figure~\ref{fig:BB}(c) and (d). We observe that the RO phenomenon can be simulated by adjusting the adversarial perturbations on training data. These experimental results support our hypothesis that the adversarial perturbations can degrade feature generalization and induce the emergence of RO.

\section{Additional Results for ROFG$_\mathrm{AS}$}
\label{D.D}
\begin{figure}[t]
\centering
    \subfigure[]{
        \includegraphics[width=0.31\columnwidth]{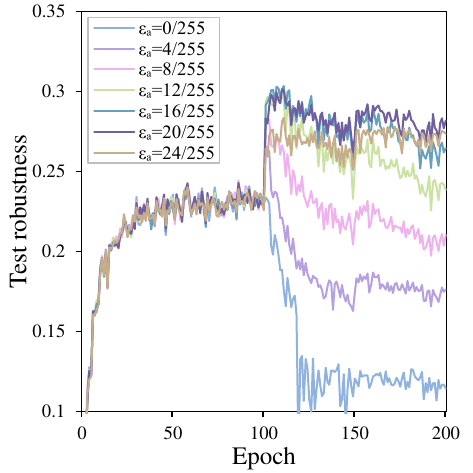}
    }
    \subfigure[]{
        \includegraphics[width=0.31\columnwidth]{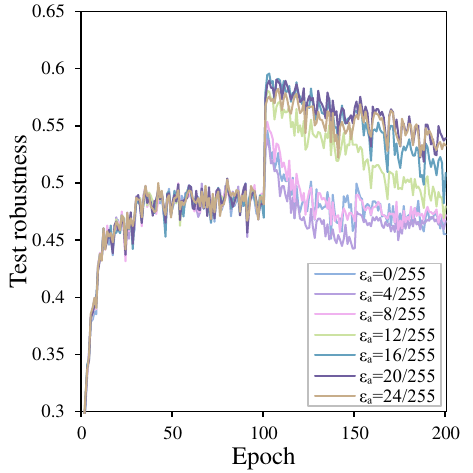}
    }
    \subfigure[]{
        \includegraphics[width=0.31\columnwidth]{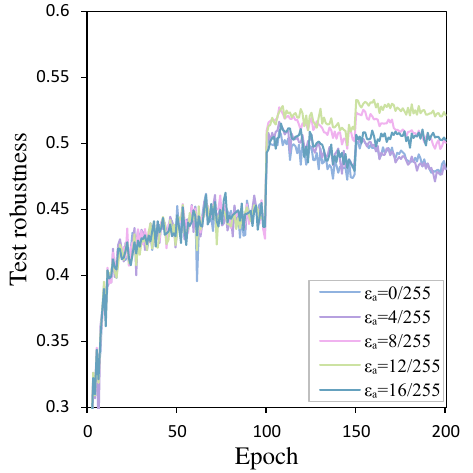}
    }
\caption{Test robustness of ROFG$_\mathrm{AS}$ (a) on CIFAR100 dataset using PreAct ResNet-18 with AT, (b) on CIFAR10 dataset using Wide ResNet-34-10 with AT, and (c) on CIFAR10 dataset using PreAct ResNet-18 with TRADES.}
\label{fig:dd}
\end{figure}

\begin{figure}[t]
\centering
    \subfigure[]{
        \includegraphics[width=0.31\columnwidth]{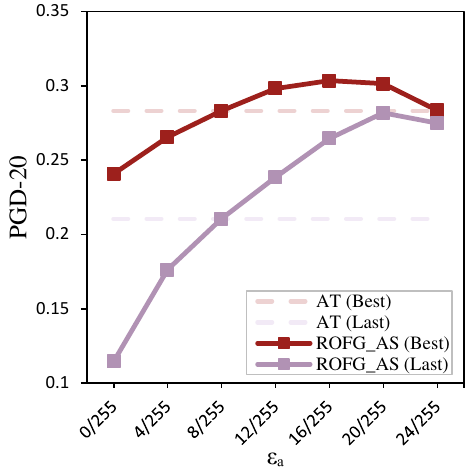}
    }
    \subfigure[]{
        \includegraphics[width=0.31\columnwidth]{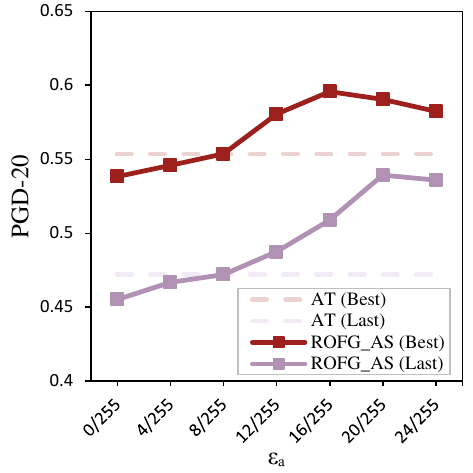}
    }
    \subfigure[]{
        \includegraphics[width=0.31\columnwidth]{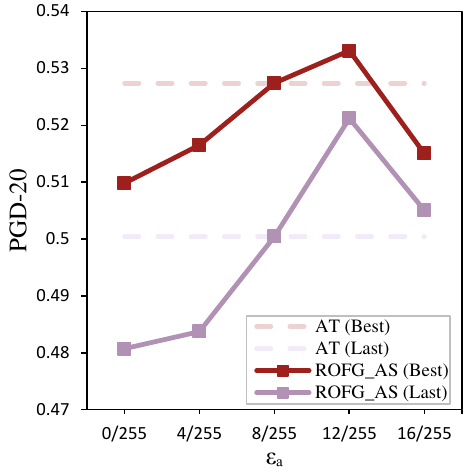}
    }
\caption{The robustness gap between the ``Best'' and ``Last'' checkpoints of ROFG$_\mathrm{AS}$ (a) on CIFAR100 dataset using PreAct ResNet-18 with AT, (b) on CIFAR10 dataset using Wide ResNet-34-10 with AT, and (c) on CIFAR10 dataset using PreAct ResNet-18 with TRADES.}
\label{fig:ddd}
\end{figure}

In this section, we present additional results for ROFG$_\mathrm{AS}$. We utilize a fixed loss threshold of 1.7 for the CIFAR10 dataset and 4.0 for the CIFAR100 dataset to distinguish small-loss training data. We conducted ROFG$_\mathrm{AS}$ with varying perturbation budgets $\epsilon_a$ across different datasets, network architectures, and AT variants. The results are summarized in Figure~\ref{fig:dd} and~\ref{fig:ddd}. Consistently, we observe a clear correlation between the applied perturbation budgets $\epsilon_a$ and the extent of RO. As the perturbation budgets $\epsilon_a$ increases, the robustness gap between the ``Best’’ and ``Last’’ models decreases. These experimental results suggest that adjusting adversarial perturbations on small-loss training data can effectively prevent the degradation of feature generalization and mitigate RO, validating our feature generalization understanding of RO.

\section{Additional Results for ROFG$_\mathrm{DA}$}
\label{E.E}
\begin{figure}[t]
\centering
    \subfigure[]{
        \includegraphics[width=0.31\columnwidth]{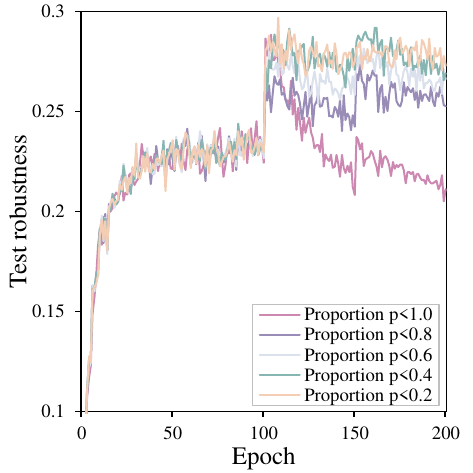}
    }
    \subfigure[]{
        \includegraphics[width=0.31\columnwidth]{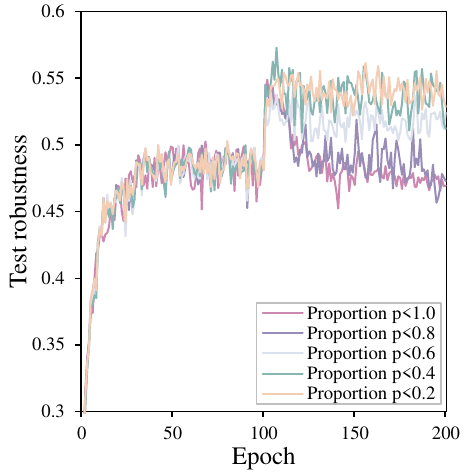}
    }
    \subfigure[]{
        \includegraphics[width=0.31\columnwidth]{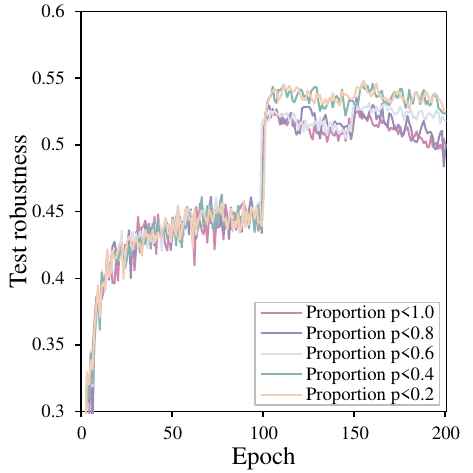}
    }
\caption{Test robustness of ROFG$_\mathrm{DA}$ (a) on CIFAR100 dataset using PreAct ResNet-18 with AT, (b) on CIFAR10 dataset using Wide ResNet-34-10 with AT, and (c) on CIFAR10 dataset using PreAct ResNet-18 with TRADES.}
\label{fig:ee}
\end{figure}

\begin{figure}[t]
\centering
    \subfigure[]{
        \includegraphics[width=0.31\columnwidth]{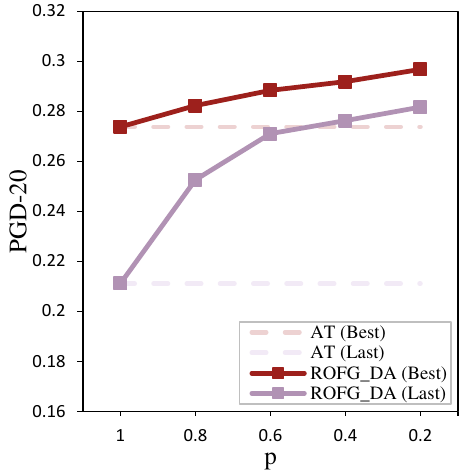}
    }
    \subfigure[]{
        \includegraphics[width=0.31\columnwidth]{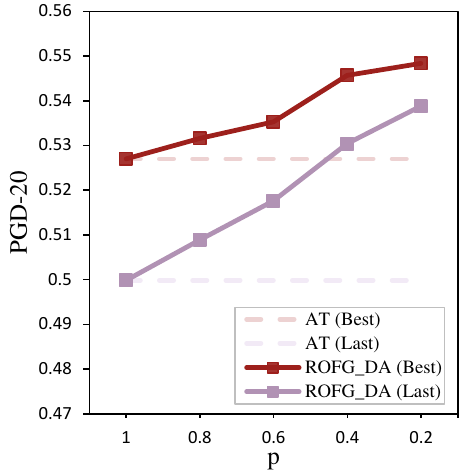}
    }
    \subfigure[]{
        \includegraphics[width=0.31\columnwidth]{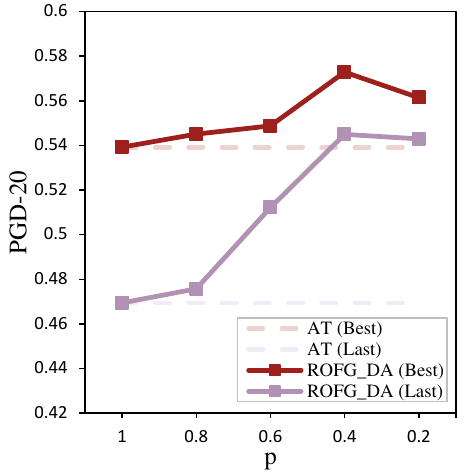}
    }
\caption{The robustness gap between the ``Best'' and ``Last'' checkpoints of ROFG$_\mathrm{DA}$ (a) on CIFAR100 dataset using PreAct ResNet-18 with AT, (b) on CIFAR10 dataset using Wide ResNet-34-10 with AT, and (c) on CIFAR10 dataset using PreAct ResNet-18 with TRADES.}
\label{fig:eee}
\end{figure}

In this section, we provide additional results of ROFG$_\mathrm{DA}$ to support our feature generalization perspective. We conducted ROFG$_\mathrm{DA}$ with different thresholds $p$ across different datasets, network architectures, and AT variants. The results are summarized in Figure~\ref{fig:ee} and~\ref{fig:eee}. Consistently, we observe a clear correlation between the threshold $p$ and the extent of RO. As the robustness gap between small-loss training data and test data decreases, the extent of RO becomes increasingly mild. These experimental results show that narrowing the model’s robustness gap between training and test data can effectively prevent the degradation of feature generalization and mitigate RO, supporting our feature generalization understanding of RO.

\section{Experimental Setup}
\label{F.F}
Our project is implemented using the PyTorch framework on servers equipped with four GeForce RTX 3080 and one Tesla V100 GPUs. The code and related models will be publicly released for verification and use. We conducted extensive experiments across different benchmark datasets (CIFAR10 and CIFAR100~\citep{krizhevsky2009learning}), network architectures (PreAct ResNet-18~\citep{he2016deep} and Wide ResNet-34-10~\citep{zagoruyko2016wide}), and adversarial training approaches (AT~\citep{madry2018towards}, TRADES~\citep{zhang2019theoretically}, AWP~\citep{wu2020adversarial}, and MLCAT~\citep{yu2022understanding}). Throughout our experiments, we utilize the infinity norm as the adversarial perturbation constraint.

For training, we follow the experimental settings outlined in Rice et al.\citep{rice2020overfitting}. The network undergoes 200 epochs of training using stochastic gradient descent (SGD) with a momentum of 0.9, weight decay of $5\times10^{-4}$, and an initial learning rate of 0.1. The learning rate is reduced by a factor of 10 at the 100th and 150th epochs. Standard data augmentation techniques, including random cropping with 4 pixels of padding and random horizontal flips, are applied.

For the adversarial setting, we use the 10-step PGD with a perturbation budget of $\epsilon=8/255$ and a step size of $\alpha=2/255$, following the standard setting of PGD-AT~\citep{madry2018towards}. In ROFG$_\mathrm{AS}$, the number of PGD steps linearly increases with the adjusted perturbation budget $\epsilon_a$. For instance, a 10-step PGD is utilized for $\epsilon_a=8/255$, while a 20-step PGD is applied for $\epsilon_a=16/255$.

For performance evaluation, we present the model's learning curves to demonstrate their effectiveness in mitigating RO. The model’s robustness is assessed using multiple attack methods, including 20-step PGD (PGD-20)~\citep{madry2018towards} and AutoAttack (AA)~\citep{croce2020reliable}. We report the results for both the ``Best’’ and ``Last’’ checkpoints, along with the standard deviations from 5 runs. The configurations for ROFG$_\mathrm{AS}$ and ROFG$_\mathrm{DA}$ are detailed in Table~\ref{hyperparameter}, and other hyperparameters for the baseline methods are set as per their original papers.

\begin{table}[t]
  \centering
  \caption{The experimental configurations of ROFG$_\mathrm{AS}$ and ROFG$_\mathrm{DA}$.}
  \label{hyperparameter}
\resizebox{\linewidth}{!}{
  \begin{tabular}{lllcccccc}
    \toprule
    \multirow{2}*{Network} & \multirow{2}*{Dataset} & \multirow{2}*{Method} & \multicolumn{3}{c}{Hyperparameter} & & \multicolumn{2}{c}{Training time per epoch}\\
    \cmidrule{4-6}
    \cmidrule{8-9}
    & & & $t$ & $\epsilon_a$ & $p$ & & GeForce RTX 3080 & Tesla V100\\
    \midrule
    \multirow{10}*{PreAct ResNet-18} & \multirow{2}*{CIFAR10} & ROFG$_{\mathrm{DA}}$ & 1.7 & - & 0.6 & & 1222.9s & -\\
     & & ROFG$_{\mathrm{AS}}$ & 1.7 & 14/255 & -& & 261.6s & -\\
    \cmidrule{2-9}
      & \multirow{2}*{CIFAR100} & ROFG$_{\mathrm{DA}}$  & 3.5 & - & 0.6 & & 1264.3s & -\\
    & & ROFG$_{\mathrm{AS}}$ & 3.5 & 15/255 & - & & 274.0s & -\\
    \cmidrule{2-9}
     & \multirow{2}*{CIFAR10} & TRADES-ROFG$_{\mathrm{DA}}$ & 1.9 & - & 0.8 & & - & -\\
    & & TRADES-ROFG$_{\mathrm{AS}}$ & 1.9 & 10/255 & -& & - & -\\
    \cmidrule{2-9}
      & \multirow{2}*{CIFAR10} & AWP-ROFG$_{\mathrm{DA}}$ & 0.8 & - & 0.8 & & 219.7s & -\\
    & & AWP-ROFG$_{\mathrm{AS}}$ & 0.8 & 12/255 & -& & 247.0s & -\\
    \cmidrule{2-9}
      & \multirow{2}*{CIFAR10} & MLCAT-ROFG$_{\mathrm{DA}}$ & 0.7 & - & 0.8& & 354.4s & -\\
    & & MLCAT-ROFG$_{\mathrm{AS}}$ & 0.7 & 11/255 & -& & 387.4s & -\\
    \midrule
    \multirow{5}*{Wide ResNet-34-10} & \multirow{1}*{CIFAR10} & ROFG$_{\mathrm{AS}}$ & 1.4 & 13/255 & -& & - & 1872.1s\\
    \cmidrule{2-9}
      & \multirow{1}*{CIFAR100} & ROFG$_{\mathrm{AS}}$ & 4.2 & 16/255 & -& & - & 2053.9s\\
    \cmidrule{2-9}
     & \multirow{1}*{CIFAR10} & TRADES-ROFG$_{\mathrm{AS}}$ & 1.9 & 10/255 & -& & - & -\\
    \cmidrule{2-9}
      & \multirow{1}*{CIFAR10} & AWP-ROFG$_{\mathrm{AS}}$ & 1.1 & 10/255 & -& & - & 1794.0s\\
    \cmidrule{2-9}
      & \multirow{1}*{CIFAR10} & MLCAT-ROFG$_{\mathrm{AS}}$ & 0.8 & 9/255 & -& & - & 2650.9s\\
    \bottomrule
  \end{tabular}
  }
\end{table}

\begin{figure}[t]
\centering
    \subfigure[AutoAugment]{
        \includegraphics[width=0.31\columnwidth]{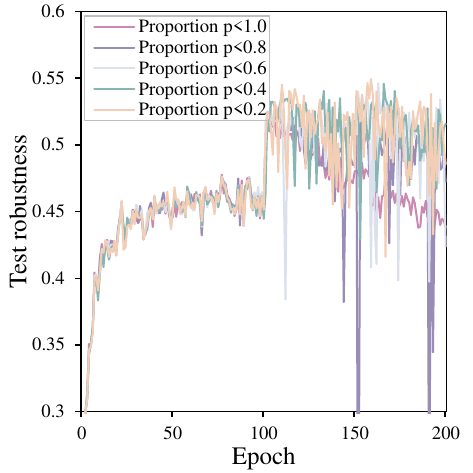}
    }
    \subfigure[RandAugment]{
        \includegraphics[width=0.31\columnwidth]{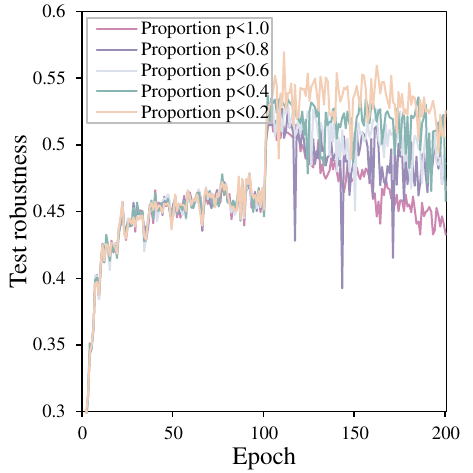}
    }
    \subfigure[TrivialAugment]{
        \includegraphics[width=0.31\columnwidth]{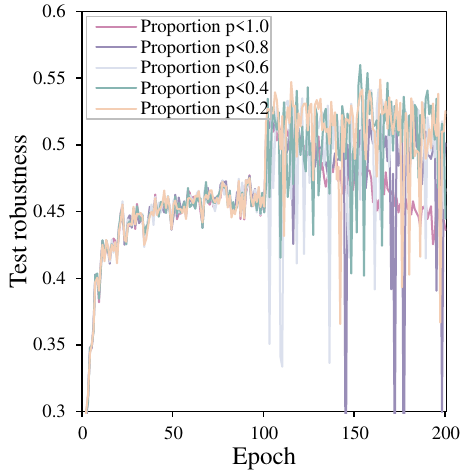}
    }
\caption{Test robustness of ROFG$_\mathrm{DA}$ with different data augmentation techniques: (a) AutoAugment; (b) RandAugment, and (c) TrivialAugment.}
\label{fig:dataaug3}
\end{figure}

\section{The Impact of Data Augmentation Techniques in ROFG$_\mathrm{DA}$}
\label{G.G}
In this section, we explore the use of different data augmentation techniques to implement ROFG$_\mathrm{DA}$. Specifically, we conduct experiments with ROFG$_\mathrm{DA}$ using three popular data augmentation techniques: AutoAugment~\citep{cubuk2019autoaugment}, RandAugment~\citep{cubuk2020randaugment}, and TrivialAugment~\citep{muller2021trivialaugment}. The results are summarized in Figure~\ref{fig:dataaug3}. We observe that the model's test robustness exhibits considerable oscillation, likely due to the significant changes in input images caused by these data augmentation techniques. Nevertheless, all implementations of ROFG$_\mathrm{DA}$ effectively mitigate RO by narrowing the robustness gap between small-loss training data and test data. These experimental results suggest that the proposed ROFG$_\mathrm{DA}$ is generally effective regardless of the chosen data augmentation techniques.


\section{Experimental Results on More Dataset and Network Architecture}
\label{H.H}
In this section, we conduct experiments with ROFG$_\mathrm{AS}$ on additional dataset and network architecture, such as VGG-16~\citep{simonyan2014very} and Tiny-ImageNet~\citep{tinyimagenet}. For the VGG-16 architecture, we set the hyperparameters of ROFG$_\mathrm{AS}$ to $\epsilon_a=12/255$ and $t=1.8$. For the Tiny-ImageNet dataset, the hyperparameters of ROFG$_\mathrm{AS}$ are set to $\epsilon_a=16/255$ and $t=7.0$. As demonstrated by the results in Table~\ref{table:4}, the proposed approach consistently mitigates RO and enhances adversarial robustness, demonstrating its effectiveness.

\begin{table*}[t]
  \small
  \centering
  \caption{Test robustbess (\%) of ROFG$_\mathrm{AS}$ on Tiny-ImageNet dataset and VGG-16 architecture.}
  \label{table:4}
\resizebox{0.99\linewidth}{!}{
  \begin{tabular}{lllcccccccc}
    \toprule
    \multirow{2}*{Method} & \multirow{2}*{Dataset} &\multirow{2}*{Network} & \multicolumn{2}{c}{Natural} && \multicolumn{2}{c}{PGD-20} && \multicolumn{2}{c}{AA}\\
    \cmidrule{4-5}
    \cmidrule{7-8}
    \cmidrule{10-11}
    & & & Best & Last && Best & Last && Best & Last \\
    \midrule
    AT & \multirow{2}*{CIFAR10} & \multirow{2}*{VGG-16} & \textbf{78.84$\pm$0.36} & \textbf{81.89$\pm$0.13} && 49.63$\pm$0.21 & 43.94$\pm$0.34 && 43.70$\pm$0.13 & 39.95$\pm$0.08 \\
    ROFG$_{\mathrm{AS}}$ & & & 75.71$\pm$0.14 & 76.93$\pm$0.38 && \textbf{53.06$\pm$0.11} & \textbf{50.62$\pm$0.09} && \textbf{46.09$\pm$0.18} & \textbf{43.87$\pm$0.15} \\
    \midrule
    AT & \multirow{2}*{Tiny-ImageNet} & \multirow{2}*{PreAct ResNet-18} & \textbf{45.51$\pm$0.54} & \textbf{45.94$\pm$0.11} && 21.34$\pm$0.25 & 15.64$\pm$0.13 && 17.10$\pm$0.10 & 13.21$\pm$0.13 \\
    ROFG$_{\mathrm{AS}}$ & & & 39.03$\pm$0.18 & 39.06$\pm$0.23 && \textbf{22.20$\pm$0.24} & \textbf{20.48$\pm$0.21} && \textbf{17.91$\pm$0.02} & \textbf{16.45$\pm$0.07}\\
    \bottomrule
  \end{tabular}
  }
\end{table*}

\end{document}